\documentclass[runningheads]{llncs}

\usepackage{eccv}

\usepackage{eccvabbrv}

\usepackage{graphicx}
\usepackage{booktabs}

\usepackage[accsupp]{axessibility}  

\usepackage{hyperref}

\usepackage{orcidlink}

\usepackage{multirow, bm}
\usepackage{algorithm}
\usepackage{algpseudocodex}

\usepackage[listings]{tcolorbox}

\usepackage{graphicx}
\makeatletter
\newcommand{\ostar}{\mathbin{\mathpalette\make@circled *}}
\newcommand{\make@circled}[2]{%
  \ooalign{$\m@th#1\smallbigcirc{#1}$\cr\hidewidth$\m@th#1#2$\hidewidth\cr}%
}
\newcommand{\smallbigcirc}[1]{%
  \vcenter{\hbox{\scalebox{0.77778}{$\m@th#1\bigcirc$}}}%
}
\makeatother

\definecolor{blue3}{HTML}{3b75af}

\definecolor{Gray}{gray}{0.94}

\newcommand{\vDelta}{{\boldsymbol \Delta}}
\newcommand{\vone}{{\bf 1}}

\newcommand{\bR}{\mathbb{R}}

\newcommand{\vf}{{\bf f}}
\newcommand{\vg}{{\bf g}}
\newcommand{\vh}{{\bf h}}

\newcommand{\vr}{{\bf r}}

\newcommand{\vx}{{\bf x}}
\newcommand{\vy}{{\bf y}}
\newcommand{\vz}{{\bf z}}

\newcommand{\vA}{{\bf A}}
\newcommand{\vB}{{\bf B}}
\newcommand{\vC}{{\bf C}}

\newcommand{\vI}{{\bf I}}

\begin{document}

\title{MTMamba: Enhancing Multi-Task Dense Scene Understanding by Mamba-Based Decoders} 

\titlerunning{MTMamba}

\author{Baijiong Lin\inst{1,4} \and
Weisen Jiang\inst{2,3} \and
Pengguang Chen\inst{5}\and
Yu Zhang\inst{3}\and
Shu Liu\inst{5,}$^\star$\and
Ying-Cong Chen\inst{1,2,4,}\thanks{Corresponding authors.}
}

\authorrunning{B.~Lin et al.}

\institute{The Hong Kong University of Science and Technology (Guangzhou) \and
The Hong Kong University of Science and Technology \and 
Southern University of Science and Technology \and 
HKUST(GZ) - SmartMore Joint Lab \and SmartMore \\
\email{\{bj.lin.email, waysonkong, akuxcw, yu.zhang.ust, liushuhust\}@gmail.com}\\
\email{yingcongchen@ust.hk}
}

\maketitle

\begin{abstract}
Multi-task dense scene understanding, which learns a model for multiple dense prediction tasks, has a wide range of application scenarios. Modeling long-range dependency and enhancing cross-task interactions are crucial to multi-task dense prediction. In this paper, we propose MTMamba, a novel Mamba-based architecture for multi-task scene understanding. It contains two types of core blocks: self-task Mamba (STM) block and cross-task Mamba (CTM) block. STM handles long-range dependency by leveraging Mamba, while CTM explicitly models task interactions to facilitate information exchange across tasks. Experiments on NYUDv2 and PASCAL-Context datasets demonstrate the superior performance of MTMamba over Transformer-based and CNN-based methods. Notably, on the PASCAL-Context dataset, MTMamba achieves improvements of +2.08, +5.01, and +4.90 over the previous best methods in the tasks of semantic segmentation, human parsing, and object boundary detection, respectively. The code is available at \url{https://github.com/EnVision-Research/MTMamba}.

\end{abstract}

\section{Introduction}
\label{sec:intro}

Multi-task dense scene understanding is an essential problem in computer vision \cite{vandenhende2021multi} and has a variety of practical applications, such as autonomous driving \cite{ishihara2021multi, liang2023multi}, healthcare \cite{hur2023genhpf}, and robotics \cite{ze2023gnfactor}. It aims to train a model for simultaneously handling multiple dense prediction tasks, such as semantic segmentation, monocular depth estimation, and surface normal estimation. 

The prevalent multi-task architecture follows an encoder-decoder framework, consisting of a task-shared encoder for feature extraction and task-specific decoders for predictions \cite{vandenhende2021multi}. This framework is very general and many variants have been proposed \cite{ye2022inverted, xu2023multi, xu2018pad, vandenhende2020mti} to improve its performance in multi-task scene understanding. One promising approach is the decoder-focused method \cite{vandenhende2021multi} with the aim of enhancing cross-task interaction in task-specific decoders through well-designed fusion modules. For example, derived from the convolutional neural network (CNN), PAD-Net \cite{xu2018pad} and MTI-Net \cite{vandenhende2020mti} incorporate a multi-modal distillation module to promote information fusion between different tasks in the decoder and achieve better performance than the encoder-decoder framework.
Since the convolution operation mainly focuses on local features \cite{bello2019attention}, recent methods \cite{ye2022inverted, xu2023multi} propose Transformer-based decoders with attention-based fusion modules. These methods leverage the attention mechanism to capture global context information, resulting in better performance than CNN-based methods. Previous works demonstrate that \textit{enhancing cross-task correlation} and \textit{modeling long-range spatial relationships} are critical for multi-task dense prediction. 

Very recently, Mamba \cite{gu2023mamba}, a new architecture derived from state space models (SSMs) \cite{gu2021combining, gu2021efficiently}, has shown better long-range dependencies modeling capacity and superior performance than Transformer models in various domains, including language modeling \cite{gu2023mamba,grazzi2024mamba,wang2024mambabyte}, graph reasoning \cite{wang2024graph,behrouz2024graph}, medical images analysis \cite{ma2024u,xing2024segmamba}, and point cloud analysis \cite{liang2024pointmamba,zhang2024point}. However, all of these works focus on single-task learning, while how to adopt Mamba for multi-task training is still under investigation. Moreover, achieving cross-task correlation in Mamba remains unexplored, which is critical for multi-task scene understanding.

To fill these gaps, in this paper, we propose \textbf{MTMamba}, a novel multi-task architecture featuring a Mamba-based decoder and superior performance in multi-task scene understanding. The overall framework is shown in Figure \ref{fig:overall_arch}. MTMamba is a decoder-focused method with two types of core blocks: the self-task Mamba (STM) block and the cross-task Mamba (CTM) block, illustrated in Figure \ref{fig:block}. Specifically, STM, inspired by Mamba, can effectively capture global context information. CTM is designed to enhance each task's features by facilitating knowledge exchange across different tasks. Therefore, through the collaboration of STM and CTM blocks in the decoder, MTMamba not only enhances cross-task interaction but also effectively handles long-range dependency.

We evaluate MTMamba on two standard multi-task dense prediction benchmark datasets, namely NYUDv2 \cite{silberman2012indoor} and PASCAL-Context \cite{chen2014detect}. Quantitative results demonstrate that MTMamba largely outperforms both CNN-based and Transformer-based methods. Notably, on the PASCAL-Context dataset, MTMamba outperforms the previous best by +2.08, +5.01, and +4.90 in semantic segmentation, human parsing, and object boundary detection tasks, respectively. Qualitative studies show that MTMamba generates better visual results with more accurate details than state-of-the-art Transformer-based methods.

Our main contributions are summarized as follows:
\begin{itemize}
\item We propose MTMamba, a novel multi-task architecture for multi-task scene understanding. It contains a novel Mamba-based decoder, which effectively models long-range spatial relationships and achieves cross-task correlation;
\item We design a novel CTM block to enhance cross-task interaction in multi-task dense prediction;
\item Experiments on two benchmark datasets demonstrate the superiority of MTMamba on multi-task dense prediction over previous CNN-based and Transformer-based methods;
\item Qualitative evaluations show that MTMamba captures discriminative features and generates precise predictions.
\end{itemize}

\section{Related Works}

\subsection{Multi-Task Learning}

Multi-task learning (MTL) is a learning paradigm that aims to simultaneously learn multiple tasks in a single model \cite{zhang2021survey}. Recent MTL research mainly focuses on multi-objective optimization \cite{linreasonable, lin2023scale, ye2021multi, ye2024first, sk18, yu2020gradient, liu2021conflict, yeadaptive} and network architecture design \cite{ye2022inverted, xu2023multi, xu2018pad, vandenhende2020mti}. In multi-task dense scene understanding, most existing works focus on designing architecture \cite{vandenhende2021multi}, especially designing specific modules in the decoder to achieve better cross-task interaction. For example, based on CNN, Xu et al. \cite{xu2018pad} introduce PAD-Net, incorporating an effective multi-modal distillation module to promote information fusion between different tasks in the decoder. MTI-Net \cite{vandenhende2020mti} is a complex multi-scale and multi-task CNN architecture with an information distillation across various feature scales. As the convolution operation mainly captures local features \cite{bello2019attention}, recent approaches \cite{ye2022inverted, xu2023multi} utilize the attention mechanism to grasp global context and develop Transformer-based decoders for multi-task scene understanding. For instance, Ye \& Xu \cite{ye2022inverted} introduce InvPT, a Transformer-based multi-task architecture, employing an effective UP-Transformer block for multi-task feature interaction at different feature scales. MQTransformer \cite{xu2023multi} designs a cross-task query attention module to enable effective task association and information exchange in the decoder.

Previous works demonstrate long-range dependency modeling and enhancing cross-task correlation are critical for multi-task dense prediction. Unlike existing methods, we propose a novel multi-task architecture derived from Mamba to capture global information better and promote cross-task interaction. 

\subsection{State Space Models}
State space models (SSMs) are a mathematical representation of dynamic systems,
which models the input-output relationship through a hidden state.
SSMs are general and have achieved great success in a wide variety of applications such as reinforcement learning~\cite{hafner2019dream},
computational neuroscience~\cite{friston2003dynamic},
and
linear dynamical systems~\cite{hespanha2018linear}.
Recently, 
SSMs are introduced as an alternative network architecture to model long-range dependency. 
Compared with CNN-based networks~\cite{krizhevsky2017imagenet, he2016deep}, which are designed for capturing local dependence,
SSMs are more powerful for long sequences;
Compared with Transformer-based networks~\cite{dosovitskiy2021an, wolf2020transformers}, which require the quadratic complexity of the sequence length,
SSMs are more computation- and memory-efficient.

Many different structures have been proposed recently to improve the expressivity and efficiency of SSMs.
Gu et al.~\cite{gu2021efficiently}
propose structured state space models (S4) to improve computational efficiency,
where the state matrix is a sum of low-rank and normal matrices.
Many follow-up works attempt to enhance the effectiveness of S4.
For example,
Fu et al.~\cite{fu2023hungry} design a new SSM layer H3 to fill the performance gap between SSMs and Transformers in language modeling.
Mehta et al.~\cite{mehta2023long}
introduce a gated state space layer using gated units for improving expressivity.
Recently, Gu \& Dao~\cite{gu2023mamba} further propose Mamba with the core operation S6,
an input-dependent selection mechanism of S4, which
achieves linear scaling in sequence length and demonstrates superior performance over Transformers on various benchmarks.
Mamba has been successfully applied in image classification~\cite{liu2024vmamba, zhu2024vision}, image segmentation~\cite{xing2024segmamba}, and graph prediction~\cite{wang2024graph}.
Different from them, which use Mamba in the single-task setting, we consider a more challenging multi-task setting and propose novel self-task and cross-task Mamba modules to capture intra-task and inter-task dependence.

\begin{figure}[!t]
\centering
\includegraphics[width=\textwidth]{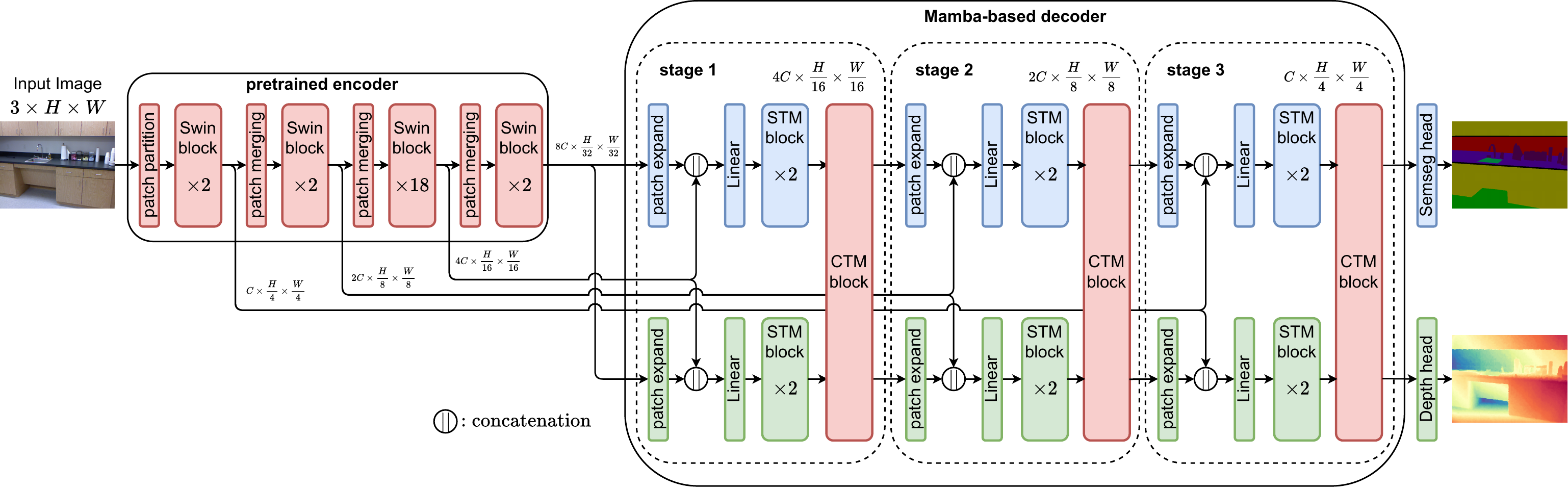}
\caption{Overview of the proposed MTMamba for multi-task dense scene understanding, illustrating with semantic segmentation (abbreviated as ``Semseg'') and depth estimation (abbreviated as ``Depth'') tasks. The red blocks are shared across all tasks, while the blue and green ones are task-specific. The pretrained encoder (Swin-Large Transformer is used) extracts multi-scale generic visual representations from the input RGB image. In the decoder, all task representations from task-specific STM blocks are fused and refined in the CTM block. Each task has its own head to generate the final predictions. Note that the structures of STM and CTM blocks (details in Figure \ref{fig:block}) in the decoder are Mamba-based (i.e., non-attention).}
\label{fig:overall_arch}
\end{figure}

\section{Methodology}

In this section, we first introduce the background knowledge of state space models and Mamba in Section \ref{sec:ssm}. Then, we introduce the overall architecture of the proposed MTMamba in Section \ref{sec:overview}. Subsequently, we delve into a detailed exploration of each part in MTMamba, including the encoder in Section \ref{sec:encoder}, the Mamba-based decoder in Section \ref{sec:mamba_decoder}, and the output head in Section \ref{sec:head}.

\subsection{Preliminaries} \label{sec:ssm}

SSMs~\cite{gu2021combining, gu2021efficiently, gu2023mamba}, 
originated from the linear systems theory~\cite{chen1984linear, hespanha2018linear},
map input sequence $x(t)\in \bR$ to output sequence $y(t)\in \bR$ though a hidden state $\vh\in\bR^N$ by a linear ordinary differential equation:
\begin{align}
\vh'(t) &= \vA \vh(t) + \vB x(t), \label{eq:ssm-1}\\ 
y(t) &= \vC^\top \vh(t) + D x(t), \label{eq:ssm-2}   
\end{align}
where $\vA \in \bR^{N\times N}$ is the state matrix,
$\vB \in \bR^{N}$ is the input matrix, 
$\vC \in \bR^{N}$ is the output matrix, 
and $D \in \bR$ is the skip connection. Equation \eqref{eq:ssm-1} defines the evolution of the hidden state $\vh(t)$, while Equation \eqref{eq:ssm-2} determines the output is composed of a linear transformation of the hidden state $\vh(t)$ and a skip connection from $x(t)$.
For the remainder of this paper, $D$ is omitted for explanation (i.e., $D=0$).

Since the continuous-time system is not suitable for digital computers and real-world data, which are usually discrete, a discretization procedure is introduced to approximate it by a discrete-time one.
Let $\Delta\in \bR$ be a discrete-time step.
Equations \eqref{eq:ssm-1} and \eqref{eq:ssm-2} are discretized as 
\begin{align}
\vh_t &= \bar{\vA}\vh_{t-1} + \bar{\vB} x_t, \label{eq:ssm-d1}  \\ 
y_t &= \bar{\vC}^\top \vh_t, 
\label{eq:ssm-d2}
\end{align}
where $x_t = x(\Delta t)$, and
\begin{align}
\bar{\vA} = \exp(\Delta \vA), \quad \bar{\vB} = (\Delta \vA)^{-1}(\exp(\Delta \vA) - \vI) \cdot \Delta \vB \approx \Delta \vB, \quad \bar{\vC} = \vC. \label{eq:discrete}
\end{align}

In S4 \cite{gu2021efficiently}, $(\vA, \vB, \vC, \Delta)$ are trainable parameters learned by gradient descent and do not explicitly depend on the input sequence, resulting in weak contextual information extraction.
To overcome this, Mamba~\cite{gu2023mamba} proposes \textbf{S6}, which introduces an input-dependent selection mechanism to allow the system to select relevant information based on the input sequence.
This is achieved by making 
$\vB$, $\vC$, and $\vDelta$ as functions of the input $x_t$. More formally, given an input sequence $\vx\in \bR^{B\times L \times C}$ where $B$ is the batch size, $L$ is the sequence length, and $C$ is the feature dimension, the input-dependent parameters $(\vB, \vC, \Delta)$ are computed as
\begin{align}
\vB &= \texttt{Linear}(\vx) \in \bR^{B\times L \times N}, \\
\vC &= \texttt{Linear}(\vx) \in \bR^{B\times L \times N}, \\
\Delta &= \texttt{SoftPlus}(\tilde{\Delta}+\texttt{Linear}(\vx)) \in \bR^{B\times L \times C},
\end{align}
where $\tilde{\Delta}\in \bR^{B\times L \times C}$ is a learnable parameter, $\texttt{SoftPlus}(\cdot)$ is the SoftPlus function, and $\texttt{Linear}(\cdot)$ is the linear layer. $\vA \in \bR^{L\times C}$ is a trainable parameter as in S4. After computing $(\vA, \vB, \vC, \Delta)$, $(\vA, \vB, \vC)$ are discretized via Equation \eqref{eq:discrete}, then the output sequence $\vy\in\bR^{B\times L \times C}$ is computed by Equations \eqref{eq:ssm-d1} and \eqref{eq:ssm-d2}.

\subsection{Overall Architecture} \label{sec:overview}

An overview of MTMamba is illustrated in Figure \ref{fig:overall_arch}. 
It contains three components: an off-the-shelf encoder, a Mamba-based decoder, and task-specific heads. 
Specifically, the encoder is shared across all tasks and responsible for extracting multi-scale generic visual representations from the input image. 
The decoder consists of three stages. Each stage contains task-specific STM blocks to capture the long-range spatial relationship for each task and a shared CTM block to enhance each task's feature by exchanging knowledge across tasks. In the end, an output head is used to generate the final prediction for each task. We introduce the details of each part as follows.

\subsection{Encoder} \label{sec:encoder}
We take the Swin Transformer \cite{liu2021swin} as an example. Consider an input RGB image $\vx\in\bR^{3\times H \times W}$, where $H$ and $W$ are the height and width of the image, respectively. The encoder employs a patch-partition module to segment the input image into non-overlapping patches. Each patch is regarded as a token, and its feature representation is a concatenation of the raw RGB pixel values. In our experiment, we use a standard patch size of $4\times 4$. Therefore, the feature dimension of each patch is $4\times 4 \times 3=48$. After patch splitting, a linear layer is applied to project the raw token into a $C$-dimensional feature embedding. 
The patch tokens, after being transformed, sequentially traverse multiple Swin Transformer blocks and patch merging layers, which collaboratively produce hierarchical feature representations. Specifically, the patch merging layer \cite{liu2021swin} is used to $2\times$ downsample the spatial dimensions (i.e., $H$ and $W$) and $2\times$ expand the feature dimension (i.e., $C$), while the Swin Transformer block focuses on learning and refining the feature representations. Formally, 
after forward passing the encoder, we obtain the output from four stages:
\begin{align}
\vf_1, \vf_2, \vf_3, \vf_4 = \texttt{encoder}(\vx),
\end{align}
where $\vf_1, \vf_2, \vf_3$, and $\vf_4$ have a size of $C\times \frac{H}{4}\times \frac{W}{4}, 2C\times \frac{H}{8}\times \frac{W}{8}, 4C\times \frac{H}{16}\times \frac{W}{16}$, and $8C\times \frac{H}{32}\times \frac{W}{32}$, respectively. 

\begin{figure}[!t]
\centering
\includegraphics[width=\textwidth]{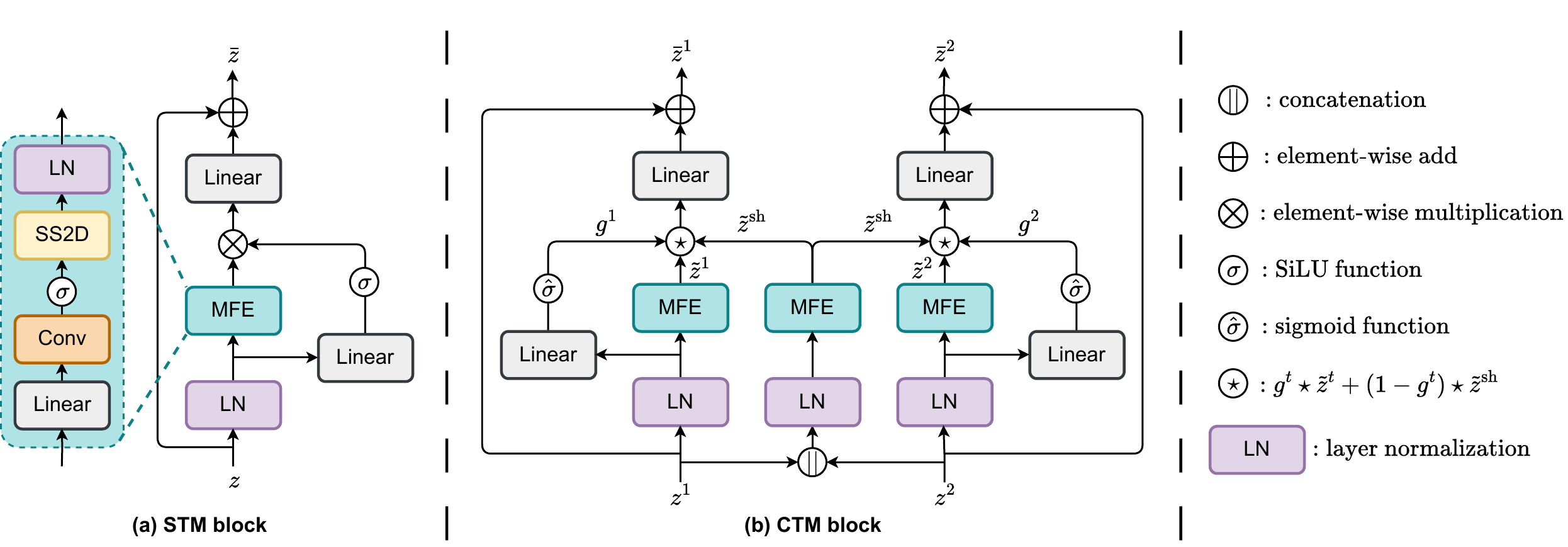}
\caption{\textbf{(a)} Illustration of the self-task Mamba (STM) block. Its core module is the Mamba-based feature extractor (MFE), where 1D S6 operation (introduced in Section \ref{sec:ssm}) is extended on 2D images, namely SS2D. MFE is responsible for learning discriminant features and an input-dependent gate $\sigma(\texttt{Linear}(\texttt{LN}(\vz)))$ further refines the learned features. \textbf{(b)} Overview of the cross-task Mamba (CTM) block, illustrating with two tasks. Suppose $T$ is the number of tasks ($T=2$ in this illustration). The CTM block inputs $T$ features, outputs $T$ features, and contains $T+1$ MFE modules. One is used to generate a global feature $\tilde{\vz}^\text{sh}$ and the other is to obtain the task-specific feature $\tilde{\vz}^t$. Each output feature is the aggregation of task-specific feature $\tilde{\vz}^t$ and global feature $\tilde{\vz}^\text{sh}$ weighted by a task-specific gate $\vg^t$. More details about these two blocks are provided in Section \ref{sec:mamba_decoder}.}
\label{fig:block}
\end{figure}

\subsection{Mamba-based Decoder} \label{sec:mamba_decoder}

\paragraph{Extend SSMs to 2D images.} Different from 1D language sequences, 2D spatial information is crucial in vision tasks. Therefore, SSMs introduced in Section \ref{sec:ssm} cannot be directly applied in 2D images. Inspired by \cite{liu2024vmamba}, we incorporate the 2D-selective-scan (SS2D) operation to address this problem. This method involves expanding image patches along four directions, generating four unique feature sequences. Then, each feature sequence is fed to an SSM (such as S6). Finally, the processed features are combined to construct the comprehensive 2D feature map. Formally, given the input feature $\vz$, the output feature $\bar{\vz}$ of SS2D is computed as
\begin{align}
\vz_v &= \texttt{expand}(\vz, v), \quad \text{for}~v \in \{1,2,3,4\}, \\
\bar{\vz}_v &= \texttt{S6}(\vz_v), \quad \text{for}~v \in \{1,2,3,4\}, \\
\bar{\vz} &= \texttt{sum}(\bar{\vz}_1, \bar{\vz}_2, \bar{\vz}_3, \bar{\vz}_4),
\end{align}
where $v\in\{1,2,3,4\}$ is the four different scanning directions, $\texttt{expand}(\vz, v)$ is to expand 2D feature map $\vz$ along direction $v$, $\texttt{S6}(\cdot)$ is the S6 operation introduced in Section \ref{sec:ssm}, and $\texttt{sum}(\cdot)$ is the element-wise add operation.

\paragraph{Mamba-based Feature Extractor (MFE).} We introduce a Mamba-based feature extractor to learn the representation of 2D images. It is a critical module in the proposed Mamba-based decoder. As shown in Figure \ref{fig:block}(a), motivated by \cite{gu2023mamba}, MFE consists of a linear layer used to expand the feature dimension by a controllable expansion factor $\alpha$, a convolution layer with an activation function for extracting local features, an SS2D operation for modeling long-range dependency, and a layer normalization to normalize the learned features. 
More formally, given the input feature $\vz$, the output $\bar{\vz}$ of MFE is calculated as
\begin{align}
\bar{\vz} = (\texttt{LN}\circ\texttt{SS2D}\circ\sigma\circ\texttt{Conv}\circ\texttt{Linear})(\vz),
\end{align}
where $\texttt{LN}(\cdot)$ is the layer normalization, $\sigma(\cdot)$ is the activation function and the SiLU function is used in our experiment, $\texttt{Conv}(\cdot)$ is the convolution operation.

\paragraph{Self-Task Mamba (STM) Block.} We introduce a self-task Mamba block for learning task-specific features based on MFE, which is illustrated in Figure \ref{fig:block}(a). Inspired by \cite{gu2023mamba}, we use an input-dependent gate to adaptively select useful representations learned from MFE. After that, a linear layer is used to reduce the feature dimension expanded in MFE. Specifically, given the input feature $\vz$, the computation in the STM block is as
\begin{align}
\vz_{\text{LN}} &= \texttt{LN}(\vz), \\
\tilde{\vz} &= \texttt{MFE}(\vz_{\text{LN}}), \\ 
\vg &= \sigma(\texttt{Linear}(\vz_{\text{LN}})), \\
\bar{\vz} &= \tilde{\vz} \star \vg, \\
\bar{\vz} &= \vz + \texttt{Linear}(\bar{\vz}),
\end{align}
where $\star$ is the element-wise multiplication.

\paragraph{Cross-Task Mamba (CTM) Block.} Although the STM block can effectively learn representations for each individual task, it lacks inter-task connections to share information which is crucial to the performance of MTL. To tackle this problem, we design a novel cross-task Mamba block (as shown in Figure \ref{fig:block}(b)) by modifying the STM block to achieve knowledge exchange across different tasks. Specifically, given all tasks' features $\{\vz^t\}_{t=1}^T$ where $T$ is the number of tasks, we first concatenate all task features and then pass it through an MFE to learn a global representation $\tilde{\vz}^\text{sh}$. Each task also learns its corresponding feature $\tilde{\vz}^t$ via its own MFE. Then, we use an input-dependent gate to aggregate the task-specific representation $\tilde{\vz}^t$ and global representation $\tilde{\vz}^\text{sh}$. Thus, each task adaptively fuses the global representation and its features. Formally, the forward process in the CTM block is as
\begin{align}
\vz_{\text{LN}}^t  &=  \texttt{LN}(\vz^t), \quad \text{for}~t\in\{1,2,\cdots,T\}, \\
\vz_{\text{LN}}^\text{sh}  &=  \texttt{LN}(\texttt{concat}(\vz^1, \vz^2,\cdots, \vz^T)), \\
\tilde{\vz}^t &=  \texttt{MFE}(\vz_{\text{LN}}^t), \quad \text{for}~t\in\{1,2,\cdots,T\}, \\
\tilde{\vz}^\text{sh} &=  \texttt{MFE}(\vz_{\text{LN}}^\text{sh}), \\
\vg^t & =  \hat{\sigma}(\texttt{Linear}(\vz_{\text{LN}}^t)), \quad \text{for}~t\in\{1,2,\cdots,T\}, \label{eq:gate} \\
\bar{\vz}^t & =  \vg^t\star\tilde{\vz}^t + (\vone-\vg^t)\star\tilde{\vz}^\text{sh}, \quad \text{for}~t\in\{1,2,\cdots,T\}, \label{eq:aggregation} \\
\bar{\vz}^t & =  \vz^t + \texttt{Linear}(\bar{\vz}^t), \quad \text{for}~t\in\{1,2,\cdots,T\},
\end{align}
where 
$\texttt{concat}(\cdot)$ is the concatenation operation, 
$\hat{\sigma}(\cdot)$ is the activation function and instead of SiLU used in STM block, we use the sigmoid function which is more suitable for generating the gating factors $\vg^t$ used in Equation \eqref{eq:aggregation}.  

\paragraph{Stage Design.} As shown in Figure \ref{fig:overall_arch}, the Mamba-based decoder contains three stages. Each stage has a similar design and comprises patch expand layers, STM blocks, and a CTM block. The patch expand layer is used to $2\times$ upsample the feature resolution and $2\times$ reduce the feature dimension. For each task, its feature will be expanded by a patch expand layer and then fused with
multi-scale features from the encoder via skip connections to complement the loss of spatial information caused by down-sampling. Then, a linear layer is used to reduce the feature dimension and two STM blocks are responsible for learning task-specific representation. Finally, a CTM block is applied to enhance each task's feature by knowledge exchange across tasks. Except for the CTM block, other modules are task-specific. More formally, the forward process of $i$-stage ($i=1,2,3$) is formulated as
\begin{align}
\vr^t_i & = \texttt{PatchExpand}(\vz^t_{i-1}) \quad \text{for}~t\in\{1,2,\cdots,T\}, \\
\vr^t_i &= \texttt{Linear}(\texttt{concat}(\vr^t_i, \vf_{4-i})), \quad \text{for}~t\in\{1,2,\cdots,T\}, \\
\vr^t_i &= \texttt{STM}(\texttt{STM}(\vr^t_i)), \quad \text{for}~t\in\{1,2,\cdots,T\}, \\
\{\vz^t_{i}\}_{t=1}^T &= \texttt{CTM}(\{\vr^t_i\}_{t=1}^T),
\end{align}
where $\vz^t_0=\vf_4$, $\texttt{PatchExpand}(\cdot)$ is the patch expand layer, $\texttt{STM}(\cdot)$ and $\texttt{CTM}(\cdot)$ are STM and CTM blocks, respectively.

\subsection{Output Head} \label{sec:head}

After obtaining each task's feature from the decoder, each task has its own output head to generate its final prediction. Inspired by \cite{cao2022swin}, each output head contains a patch expand layer and a linear layer, which is lightweight. Specifically, given the $t$-th task feature $\vz^t$ with the size of $C\times \frac{H}{4}\times \frac{W}{4}$ from the decoder, the patch expand layer performs $4\times$ up-sampling to restore the resolution of the
feature maps to the input resolution $H \times W$, and then the linear layer is used to output the final pixel-wise prediction.

\section{Experiments}
In this section, we conduct extensive experiments to demonstrate the effectiveness of the proposed MTMamba in multi-task dense scene understanding. 

\subsection{Experimental Setups}
\paragraph{Datasets.} Following \cite{ye2022inverted,xu2023multi}, experiments are conducted on two benchmark datasets with multi-task labels: NYUDv2 \cite{silberman2012indoor} and PASCAL-Context \cite{chen2014detect}. The NYUDv2 dataset comprises a variety of indoor scenes, containing 795 and 654 RGB images for training and testing, respectively. It consists of four tasks: $40$-class semantic segmentation (Semseg), monocular depth estimation (Depth), surface normal estimation (Normal), and object boundary detection (Boundary). 
The PASCAL-Context dataset, derived from the PASCAL dataset \cite{everingham2010pascal}, includes both indoor and outdoor scenes and provides pixel-wise labels for tasks like semantic segmentation, human parsing (Parsing), and object boundary detection, with additional labels for surface normal estimation and saliency detection tasks generated by \cite{maninis2019attentive}. It contains 4,998 training images and 5,105 testing images.

\paragraph{Implementation Details.} We use Swin-Large Transformer \cite{liu2021swin} pretrained on the ImageNet-22K dataset \cite{deng2009imagenet} as the encoder. All models are trained with a batch size of 8 for 50,000 iterations. The AdamW optimizer \cite{loshchilov2018decoupled} is adopted with a learning rate of $10^{-4}$ and a weight decay of $10^{-5}$. The polynomial learning rate scheduler is used in the training process. The expansion factor $\alpha$ in MFE is set to $2$. Following \cite{ye2022inverted}, we resize the input images of NYUDv2 and PASCAL-Context as $448\times 576$ and $512\times 512$, respectively, and use the same data augmentation including random color jittering, random cropping, random scaling, and random horizontal flipping. We use $\ell_1$ loss for depth estimation and surface normal estimation tasks and the cross-entropy loss for other tasks.

\begin{table}[!t]
\centering
\tabcolsep=0.08cm
\caption{Comparison with state-of-the-art methods on NYUDv2 (\textbf{left}) and PASCAL-Context (\textbf{right}) datasets. $\uparrow (\downarrow)$ indicates that a higher (lower) result corresponds to better performance.  The best and second best results are highlighted in \textbf{bold} and \underline{underline}, respectively.}
\resizebox{0.448\textwidth}{!}{
\begin{tabular}{lcccc}
\toprule
\multirow{2}{*}{\textbf{Method}} & \textbf{Semseg} & \textbf{Depth} & \textbf{Normal} & \textbf{Boundary}\\
& mIoU$\uparrow$ & RMSE$\downarrow$ & mErr$\downarrow$ & odsF$\uparrow$\\
\midrule
\multicolumn{5}{c}{\textit{CNN-based decoder}} \\
Cross-Stitch \cite{misra2016cross} & 36.34 & 0.6290 & 20.88 & 76.38\\
PAP \cite{zhang2019pattern} & 36.72 & 0.6178 & 20.82 & 76.42\\
PSD \cite{zhou2020pattern} & 36.69 & 0.6246 & 20.87 & 76.42 \\
PAD-Net \cite{xu2018pad} & 36.61 & 0.6270 & 20.85 & 76.38 \\
MTI-Net \cite{vandenhende2020mti} & 45.97 & 0.5365 & 20.27 & 77.86 \\
ATRC \cite{bruggemann2021exploring} & 46.33 & 0.5363 & 20.18 & 77.94 \\
\midrule
\multicolumn{5}{c}{\textit{Transformer-based decoder}} \\
InvPT \cite{ye2022inverted} & 53.56 & \underline{0.5183} & \underline{19.04} & 78.10\\
MQTransformer \cite{xu2023multi} & \underline{54.84} & 0.5325 & 19.67 & \underline{78.20}\\
\midrule
\multicolumn{5}{c}{\textit{Mamba-based decoder}} \\
MTMamba (\textbf{ours})  & \textbf{55.82} & \textbf{0.5066} & \textbf{18.63} & \textbf{78.70}\\
\bottomrule
\end{tabular}}
\hfill
\resizebox{0.532\textwidth}{!}{
\begin{tabular}{lccccc}
\toprule
\multirow{2}{*}{\textbf{Method}} & \textbf{Semseg} & \textbf{Parsing} & \textbf{Saliency} & \textbf{Normal} & \textbf{Boundary}\\
& mIoU$\uparrow$ & mIoU$\uparrow$ & maxF$\uparrow$ & mErr$\downarrow$ & odsF$\uparrow$\\
\midrule
\multicolumn{6}{c}{\textit{CNN-based decoder}} \\
ASTMT \cite{maninis2019attentive} & 68.00 & 61.10 & 65.70 & 14.70 & 72.40\\
PAD-Net \cite{xu2018pad} &  53.60 & 59.60 & 65.80 & 15.30 & 72.50 \\
MTI-Net \cite{vandenhende2020mti}  & 61.70 & 60.18 & 84.78 & 14.23 & 70.80 \\
ATRC \cite{bruggemann2021exploring}  & 62.69 & 59.42 & 84.70 & 14.20 & 70.96 \\
ATRC-ASPP \cite{bruggemann2021exploring} & 63.60 & 60.23 & 83.91 & 14.30 & 70.86 \\
ATRC-BMTAS \cite{bruggemann2021exploring} & 67.67 & 62.93 & 82.29 & 14.24 & 72.42 \\
\midrule
\multicolumn{6}{c}{\textit{Transformer-based decoder}} \\
InvPT \cite{ye2022inverted} & \underline{79.03} & \underline{67.61} & \textbf{84.81} & \underline{14.15} & 73.00\\
MQTransformer \cite{xu2023multi} & 78.93 & 67.41 & 83.58 & 14.21 & \underline{73.90}\\
\midrule
\multicolumn{6}{c}{\textit{Mamba-based decoder}} \\
MTMamba (\textbf{ours}) & \textbf{81.11} & \textbf{72.62} & \underline{84.14} & \textbf{14.14} & \textbf{78.80}\\
\bottomrule
\end{tabular}}

\label{tab:nyudv2}
\end{table}

\paragraph{Evaluation Metrics.} Following \cite{ye2022inverted}, we use mean intersection over union (mIoU) for semantic segmentation and human parsing tasks, root mean square error (RMSE) for monocular depth estimation task, mean error (mErr) for surface normal estimation task, maximal F-measure (maxF) for saliency detection task, and optimal-dataset-scale F-measure (odsF) for object boundary detection task. Besides, we use the average relative MTL performance $\Delta_m$ (defined in \cite{vandenhende2021multi}) as the overall performance metric. 

\subsection{Comparison with State-of-the-art Methods}

We compare the proposed MTMamba method with two types of MTL methods: CNN-based methods, including Cross-Stitch \cite{misra2016cross}, PAP \cite{zhang2019pattern}, PSD \cite{zhou2020pattern}, PAD-Net \cite{xu2018pad}, MTI-Net \cite{vandenhende2020mti}, ATRC \cite{bruggemann2021exploring}, and ASTMT \cite{maninis2019attentive}, and Transformer-based methods, i.e., InvPT \cite{ye2022inverted} and MQTransformer \cite{xu2023multi}.

\begin{figure}[!t]
\centering
\includegraphics[width=.76\textwidth]{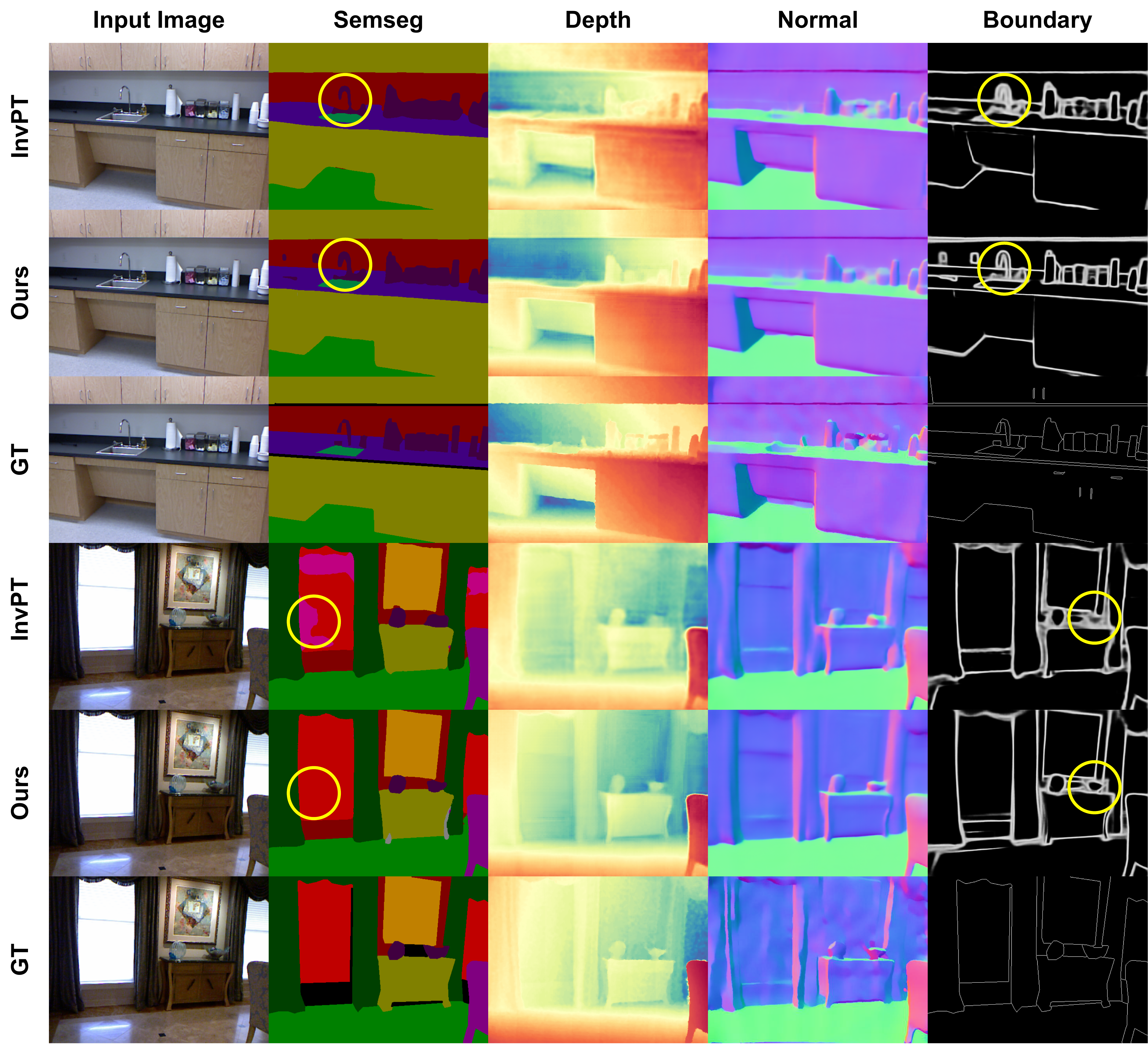}
\caption{Qualitative comparison with state-of-the-art method (i.e., InvPT \cite{ye2022inverted}) on the NYUDv2 dataset. The proposed method generates better predictions with more accurate details as marked in yellow circles. Zoom in for more details.}
\label{fig:qualitative_nyu}
\end{figure}

Table \ref{tab:nyudv2} shows the results on NYUDv2 and PASCAL-Context datasets. As can be seen, MTMamba shows superior performance in all four tasks on NYUDv2. For example, the performance of the semantic segmentation task has notably improved from the Transformer-based methods (i.e., InvPT and MQTransformer), increasing by +2.26 and +0.98, respectively, which demonstrates the effectiveness of MTMamba. 
The results on PASCAL-Context show the clear superiority of MTMamba. Especially, MTMamba significantly improves the previous best by +2.08, +5.01, and +4.90 in semantic segmentation, human parsing, and object boundary detection tasks, respectively, showing the effectiveness of MTMamba again. The qualitative comparison with InvPT on NYUDv2 and PASCAL-Context is shown in Figures \ref{fig:qualitative_nyu} and \ref{fig:qualitative_pascal}, showing that MTMmaba provides more precise predictions and details.

\begin{table}[!t]
\centering
\tabcolsep=0.08cm
\caption{Effectiveness of the STM and CTM blocks on NYUDv2. Swin-Large encoder is used in this experiment. ``Multi-task'' denotes an MTL model where each task only uses two standard Swin Transformer blocks in each decoder stage. ``Single-task'' is the single-task counterpart of ``Multi-task''. $\blacklozenge$, $\spadesuit$, $\blacksquare$, and $\bigstar$ are different variants of MTMamba. $\bigstar$ is the default configuration of MTMamba. $\uparrow (\downarrow)$ indicates that a higher (lower) result corresponds to better performance. }
\resizebox{.84\textwidth}{!}{
\begin{tabular}{lc|ccccc|cc}
\toprule
\multirow{2}{*}{\textbf{Method}} & \textbf{Each Decoder} 
& \textbf{Semseg} & \textbf{Depth} & \textbf{Normal} & \textbf{Boundary} & \bm{$\Delta_m$}[\%] & \textbf{\#Param} & \textbf{FLOPs}\\
& \textbf{Stage} & mIoU$\uparrow$ & RMSE$\downarrow$ & mErr$\downarrow$ & odsF$\uparrow$ & $\uparrow$ & MB$\downarrow$ & GB$\downarrow$\\
\midrule
Single-task & 2*Swin & 54.32 & 0.5166 & 19.21 & 77.30 & 0.00 & 888.77 & 1074.79\\
Multi-task & 2*Swin & 53.72 & 0.5239 & 19.97 & 76.50 & -1.87 & 303.18 & 466.35\\
\midrule
\multirow{4}{*}{MTMamba} & $^\blacklozenge$1*STM & 54.61 & 0.5059 & 19.00 & 77.40 & +0.95 & 252.51 & 354.13\\
& $^\spadesuit$2*STM & 54.66 & \textbf{0.4984} & 18.81 & 78.20 & +1.84 & 276.48 & 435.47\\
& $^\blacksquare$3*STM & 54.75 & 0.5054 & 18.81 & 78.20 & +1.55 & 300.45 & 516.82\\
& $^\bigstar$2*STM+1*CTM & \textbf{55.82} & {0.5066} & \textbf{18.63} & \textbf{78.70} & \textbf{+2.38} & 307.99 & 540.81\\
\bottomrule
\end{tabular}}
\label{tab:nyudv2_ablation}
\end{table}

\begin{table}[!t]
\centering
\tabcolsep=0.08cm
\caption{Effectiveness of MFE module in MTMamba on NYUDv2. Swin-Large encoder is used. ``W-MSA'' is the window-based multi-head self-attention module in Swin Transformer \cite{liu2021swin}. ``MFE'' denotes all MFE modules in both STM and CTM blocks.}
\resizebox{.76\linewidth}{!}{
\begin{tabular}{lccccc|cc}
\toprule
& \textbf{Semseg} & \textbf{Depth} & \textbf{Normal} & \textbf{Boundary} & \multirow{1}{*}{\bm{$\Delta_m$}[\%]} & \textbf{\#Param} & \textbf{FLOPs}\\
& mIoU$\uparrow$ & RMSE$\downarrow$ & mErr$\downarrow$ & odsF$\uparrow$ & $\uparrow$ & MB$\downarrow$ & GB$\downarrow$\\
\midrule
MFE$\rightarrow$W-MSA & 54.57 & 0.5109 & 19.95 & 76.60 & -0.79 & 451.81 & 884.61 \\
MTMamba & \textbf{55.82} & \textbf{0.5066} & \textbf{18.63} & \textbf{78.70} & \textbf{+2.38} & 307.99 & 540.81 \\
\bottomrule
\end{tabular}}
\label{tab:mfe}
\end{table}

\begin{table}[!t]
\centering
\tabcolsep=0.08cm
\caption{Effectiveness of linear gate in MTMamba on NYUDv2. Swin-Large encoder is used. ``W-MSA'' is the window-based multi-head self-attention module in Swin Transformer \cite{liu2021swin}. ``Linear'' denotes all linear gates in both STM and CTM blocks.}
\resizebox{.76\linewidth}{!}{
\begin{tabular}{lccccc|cc}
\toprule
& \textbf{Semseg} & \textbf{Depth} & \textbf{Normal} & \textbf{Boundary} & \multirow{1}{*}{\bm{$\Delta_m$}[\%]} & \textbf{\#Param} & \textbf{FLOPs}\\
& mIoU$\uparrow$ & RMSE$\downarrow$ & mErr$\downarrow$ & odsF$\uparrow$ & $\uparrow$ & MB$\downarrow$ & GB$\downarrow$\\
\midrule
Linear$\rightarrow$W-MSA & 55.01 & \textbf{0.4990} & 18.73 & 78.20 & +2.08 & 345.33 & 659.29\\
MTMamba & \textbf{55.82} & {0.5066} & \textbf{18.63} & \textbf{78.70} & \textbf{+2.38} & 307.99 & 540.81 \\
\bottomrule
\end{tabular}}
\label{tab:linear_gate}
\end{table}

\subsection{Model Analysis}
\paragraph{Effectiveness of STM and CTM Blocks.} The decoder of MTMamba contains two types of core blocks: STM and CTM blocks. 
We experiment on NYUDv2 to study the effectiveness of each type when the encoder is fixed as a Swin-Large Transformer. The results are shown in Table \ref{tab:nyudv2_ablation},
where ``Multi-task'' represents an MTL model using two standard Swin Transformer blocks in each decoder stage for each task, and ``Single-task'' is the single-task counterpart of ``Multi-task'' (i.e., each task has a task-specific encoder-decoder). 
According to Table \ref{tab:nyudv2_ablation}, the STM block achieves better performance and is more efficient than the Swin Transformer block ($\spadesuit$ vs. ``Multi-task''), demonstrating that Mamba is more beneficial to multi-task dense prediction. 
Simply increasing the number of STM blocks from two to three fails to boost the performance.
However, when the CTM is used, 
MTMamba has a significantly better performance in terms of $\Delta_m$
($\bigstar$ vs. $\spadesuit$/$\blacksquare$).
Moreover, the default configuration of MTMamba (i.e., $\bigstar$) significantly outperforms ``Single-task'' on all tasks, showing that MTMamba is more powerful.

\begin{table}[!t]
\begin{minipage}{.5\linewidth}
\centering
\tabcolsep=0.08cm
\caption{Effectiveness of cross-task interaction in CTM block, i.e., Equation \eqref{eq:aggregation}, on the NYUDv2 dataset. Swin-Large encoder is used in this experiment. ``adaptive $\vg^t$'' means that $\vg^t$ is computed by Equation \eqref{eq:gate}. $\uparrow (\downarrow)$ indicates that a higher (lower) result corresponds to better performance.}
\resizebox{\linewidth}{!}{
\begin{tabular}{lccccc}
\toprule
& \textbf{Semseg} & \textbf{Depth} & \textbf{Normal} & \textbf{Boundary} & \bm{$\Delta_m$}[\%]\\
& mIoU$\uparrow$ & RMSE$\downarrow$ & mErr$\downarrow$ & odsF$\uparrow$ & $\uparrow$\\
\midrule
\quad $\vg^t=0$ & 55.37 & 0.5087 & 18.76 & 78.30 & +1.77\\
\quad $\vg^t=1$ & 54.50 & \textbf{0.4981} & 18.83 & 78.20 & +1.76\\
adaptive $\vg^t$ & \textbf{55.82} & {0.5066} & \textbf{18.63} & \textbf{78.70} & \textbf{+2.38}\\
\bottomrule
\end{tabular}}
\label{tab:nyudv2_ablation_ctm}
\end{minipage}\hfill
\begin{minipage}{.46\linewidth}
\centering
\tabcolsep=0.08cm
\caption{Performance of MTMamba with different scales of Swin Transformer encoder on the NYUDv2 dataset. $\uparrow (\downarrow)$ indicates that a higher (lower) result corresponds to better performance.}
\resizebox{\linewidth}{!}{
\begin{tabular}{lcccc}
\toprule
\multirow{2}{*}{\textbf{Encoder}} 
& \textbf{Semseg} & \textbf{Depth} & \textbf{Normal} & \textbf{Boundary}\\
& mIoU$\uparrow$ & RMSE$\downarrow$ & mErr$\downarrow$ & odsF$\uparrow$ \\
\midrule
Swin-Tiny & 49.25 & 0.5299 & 19.74 & 76.90 \\
Swin-Small & 51.93 & 0.5246 & 19.45 & 77.80 \\
Swin-Base & 53.62 & 0.5126 & 19.28 & 77.70\\
Swin-Large & \textbf{55.82} & \textbf{0.5066} & \textbf{18.63} & \textbf{78.70} \\
\bottomrule
\end{tabular}}
\label{tab:backbone}
\end{minipage}
\end{table}

\paragraph{Effectiveness of MFE Module.} As shown in Figure \ref{fig:block}, the MFE module is SSM-based and is the core of both STM and CTM blocks. We conduct an experiment by replacing all MFE modules in MTMamba with the attention module. As shown in Table \ref{tab:mfe}, MFE is more effective and efficient than attention. 

\paragraph{Effectiveness of Linear Gate.} As shown in Figure \ref{fig:block}, in both STM and CTM blocks, we use an input-dependent gate to select useful representations adaptively from MFE modules. The linear layer is a simple but effective option for the gate function. We conduct an experiment by replacing all linear gates in MTMamba with the attention-based gate on the NYUDv2 dataset. As shown in Table \ref{tab:linear_gate}, the linear gate (i.e., MTMamba) performs comparably to the attention gate in terms of $\Delta_m$, while the linear gate is more efficient.

\paragraph{Effectiveness of Cross-task Interaction in CTM Block.} The core of the CTM block is the cross-task interaction, i.e., Equation \eqref{eq:aggregation}, where we fuse task-specific representation $\tilde{\vz}^t$ and shared representation $\tilde{\vz}^\text{sh}$ via a task-specific gate $\vg^t$. In this experiment, we study its effectiveness by comparing it with the cases of $\vg^t=0$ and $\vg^t=1$. The experiments are conducted with a Swin-Large Transformer encoder on NYUDv2. The results are shown in Table \ref{tab:nyudv2_ablation_ctm}. As can be seen, using a specific $\tilde{\vz}^t$ (i.e., the case of $\vg^t=0$) or shared $\tilde{\vz}^\text{sh}$ (i.e., the case of $\vg^t=1$) degrades the performance, demonstrating that the adaptive fusion is better.

\paragraph{Performance on Different Encoders.} In this experiment, we investigate the performance of the proposed MTMamba with different scales of Swin Transformer encoder on the NYUDv2 dataset. The results are shown in Table \ref{tab:backbone}. As can be seen, as the model capacity increases, all the tasks perform better accordingly. 

\begin{figure}[!t]
\centering
\includegraphics[width=.85\textwidth]{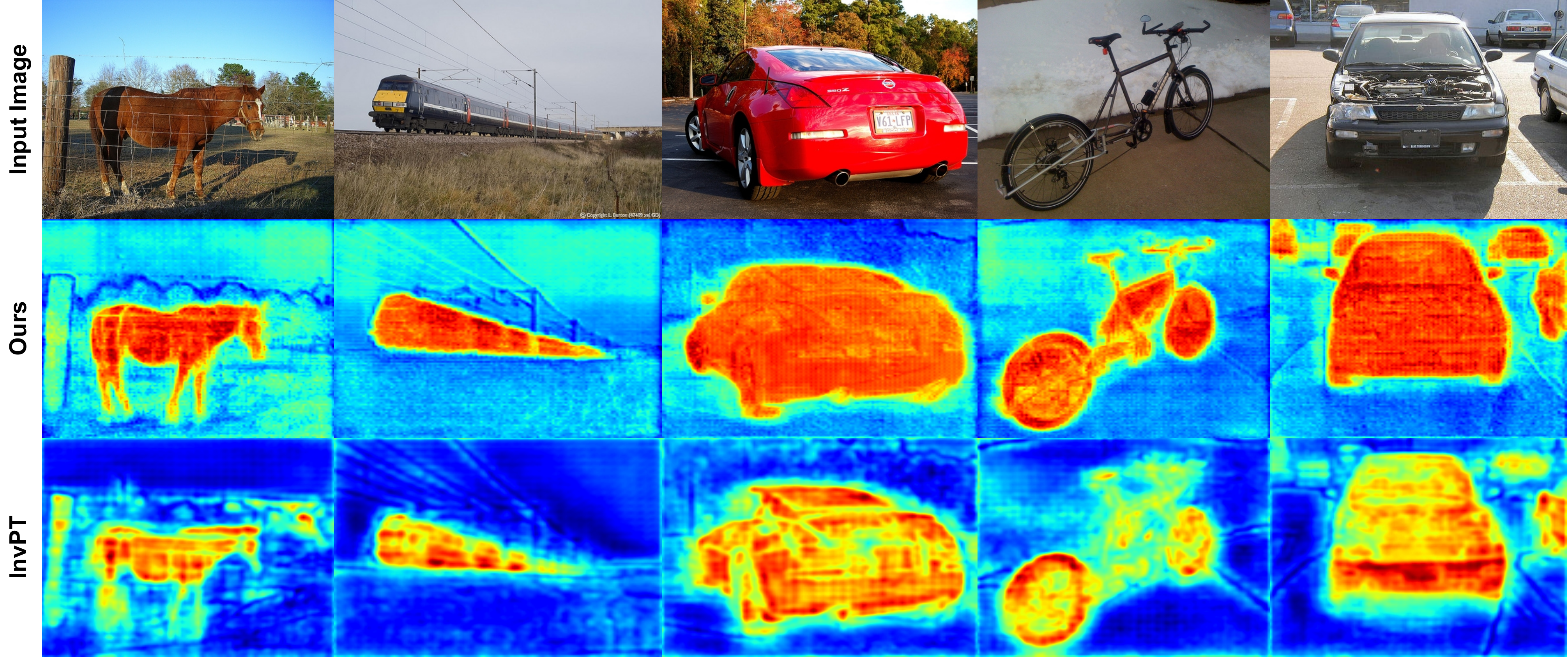}
\caption{Visualization of the final decoder feature of semantic segmentation. Compared with InvPT \cite{ye2022inverted}, our method generates more discriminative features.}
\label{fig:qualitative_feature}
\end{figure}

\subsection{Qualitative Evaluations}

\paragraph{Visualization of Learned Features.} Figure \ref{fig:qualitative_feature} shows the comparison of the final decoder feature between MTMamba and Transformer-based method InvPT \cite{ye2022inverted} in the semantic segmentation task. As can be seen, our method highly activates the regions with contextual and semantic information, which means it captures more discriminative features, resulting in better segmentation performance.

\begin{figure}[!t]
\centering
\includegraphics[width=.82\textwidth]{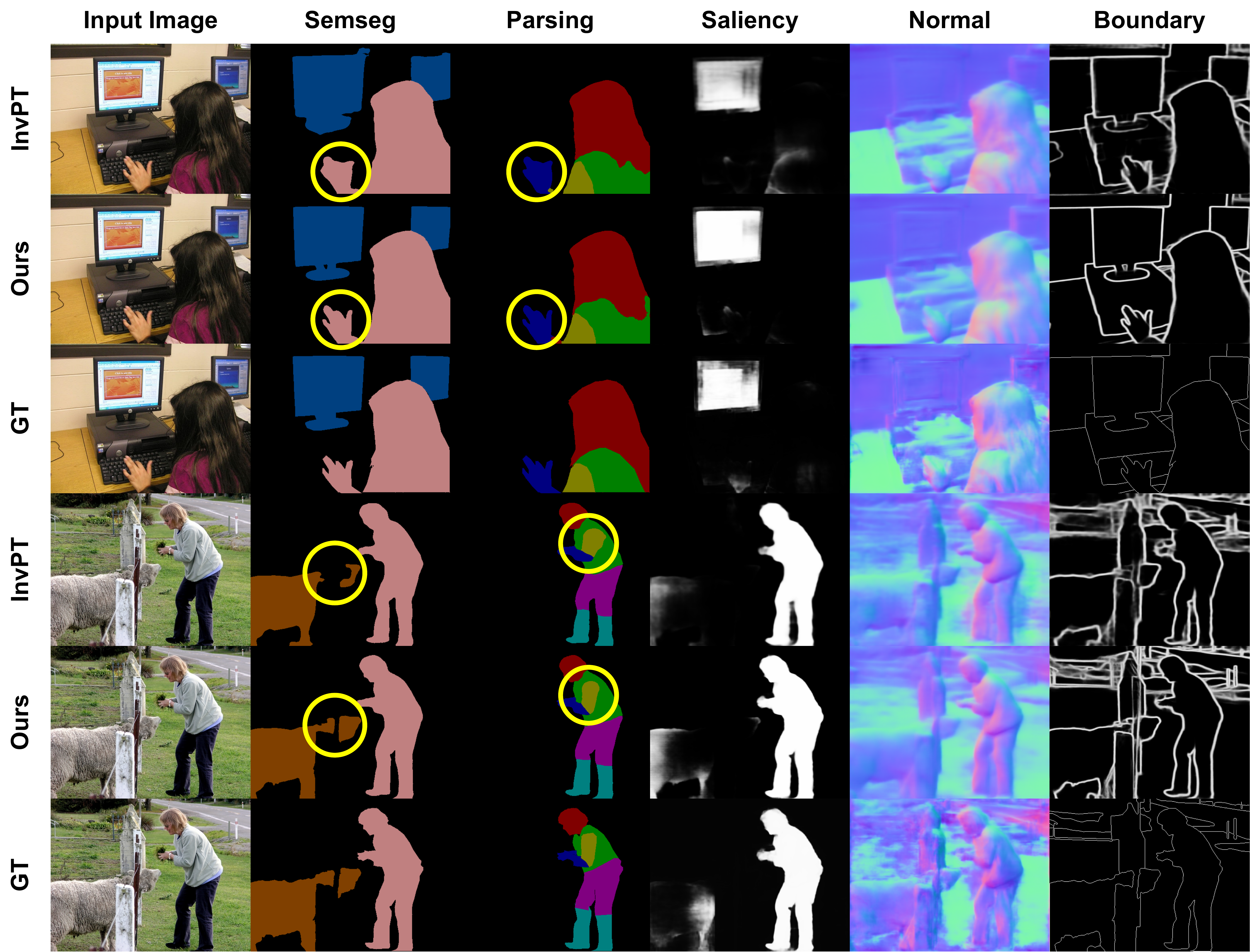}
\caption{Qualitative comparison with state-of-the-art method (i.e., InvPT \cite{ye2022inverted}) on the PASCAL-Context dataset. The proposed method generates better predictions with more accurate details as marked in yellow circles. Zoom in for more details.}
\label{fig:qualitative_pascal}
\end{figure}

\paragraph{Visualization of Predictions.}
We conduct qualitative studies by comparing the output predictions from our proposed MTMamba against the state-of-the-art Transformer-based method, InvPT \cite{ye2022inverted}. Figures \ref{fig:qualitative_nyu} and \ref{fig:qualitative_pascal} show the qualitative results on the NYUDv2 and PASCAL-Context datasets, respectively. As can be seen, our method has better visual results than InvPT in all tasks. For example, as highlighted with yellow circles in Figure \ref{fig:qualitative_nyu}, MTMamba generates more accurate results with better alignments for the semantic segmentation task and clearer object boundaries for the object boundary detection task. Figure \ref{fig:qualitative_pascal} demonstrates that MTMamba produces better predictions with more accurate details (like the fingers as highlighted) for both semantic segmentation and human parsing tasks and more distinct boundaries for the object boundary detection task. Hence, both qualitative study (Figures \ref{fig:qualitative_nyu} and \ref{fig:qualitative_pascal}) and quantitative study (Table \ref{tab:nyudv2}) show the superior performance of the proposed MTMamba method.

\section{Conclusion}
In this paper, we propose MTMamba, a novel multi-task architecture with a Mamba-based decoder for multi-task dense scene understanding. With two core blocks (STM and CTM blocks), MTMamba can effectively model long-range dependency and achieve cross-task interaction. Experiments on two benchmark datasets demonstrate that the proposed MTMamba achieves better performance than previous CNN-based and Transformer-based methods.

\section*{Acknowledgements}
This work is supported by Guangzhou-HKUST(GZ) Joint Funding Scheme (No. 2024A03J0241).


%
%
\bibliographystyle{splncs04}
\bibliography{main}

\begin{thebibliography}{10}
\providecommand{\url}[1]{\texttt{#1}}
\providecommand{\urlprefix}{URL }
\providecommand{\doi}[1]{https://doi.org/#1}

\bibitem{behrouz2024graph}
Behrouz, A., Hashemi, F.: {Graph Mamba}: Towards learning on graphs with state
  space models. arXiv preprint arXiv:2402.08678  (2024)

\bibitem{bello2019attention}
Bello, I., Zoph, B., Vaswani, A., Shlens, J., Le, Q.V.: Attention augmented
  convolutional networks. In: IEEE/CVF International Conference on Computer
  Vision (2019)

\bibitem{bruggemann2021exploring}
Br{\"u}ggemann, D., Kanakis, M., Obukhov, A., Georgoulis, S., Van~Gool, L.:
  Exploring relational context for multi-task dense prediction. In: IEEE/CVF
  International Conference on Computer Vision (2021)

\bibitem{cao2022swin}
Cao, H., Wang, Y., Chen, J., Jiang, D., Zhang, X., Tian, Q., Wang, M.:
  Swin-unet: Unet-like pure transformer for medical image segmentation. In:
  European Conference on Computer Vision (2022)

\bibitem{chen1984linear}
Chen, C.T.: Linear system theory and design. Saunders college publishing (1984)

\bibitem{chen2014detect}
Chen, X., Mottaghi, R., Liu, X., Fidler, S., Urtasun, R., Yuille, A.: Detect
  what you can: Detecting and representing objects using holistic models and
  body parts. In: IEEE Conference on Computer Vision and Pattern Recognition
  (2014)

\bibitem{deng2009imagenet}
Deng, J., Dong, W., Socher, R., Li, L.J., Li, K., Fei-Fei, L.: Imagenet: A
  large-scale hierarchical image database. In: IEEE Conference on Computer
  Vision and Pattern Recognition (2009)

\bibitem{dosovitskiy2021an}
Dosovitskiy, A., Beyer, L., Kolesnikov, A., Weissenborn, D., Zhai, X.,
  Unterthiner, T., Dehghani, M., Minderer, M., Heigold, G., Gelly, S.,
  Uszkoreit, J., Houlsby, N.: An image is worth 16x16 words: Transformers for
  image recognition at scale. In: International Conference on Learning
  Representations (2021)

\bibitem{everingham2010pascal}
Everingham, M., Van~Gool, L., Williams, C.K., Winn, J., Zisserman, A.: The
  pascal visual object classes (voc) challenge. International Journal of
  Computer Vision  \textbf{88},  303--338 (2010)

\bibitem{friston2003dynamic}
Friston, K.J., Harrison, L., Penny, W.: Dynamic causal modelling. Neuroimage
  (2003)

\bibitem{fu2023hungry}
Fu, D.Y., Dao, T., Saab, K.K., Thomas, A.W., Rudra, A., Re, C.: {Hungry Hungry
  Hippos}: Towards language modeling with state space models. In: International
  Conference on Learning Representations (2023)

\bibitem{grazzi2024mamba}
Grazzi, R., Siems, J., Schrodi, S., Brox, T., Hutter, F.: Is mamba capable of
  in-context learning? arXiv preprint arXiv:2402.03170  (2024)

\bibitem{gu2023mamba}
Gu, A., Dao, T.: Mamba: Linear-time sequence modeling with selective state
  spaces. arXiv preprint arXiv:2312.00752  (2023)

\bibitem{gu2021efficiently}
Gu, A., Goel, K., Re, C.: Efficiently modeling long sequences with structured
  state spaces. In: International Conference on Learning Representations (2022)

\bibitem{gu2021combining}
Gu, A., Johnson, I., Goel, K., Saab, K., Dao, T., Rudra, A., R{\'e}, C.:
  Combining recurrent, convolutional, and continuous-time models with linear
  state space layers. In: Neural Information Processing Systems (2021)

\bibitem{hafner2019dream}
Hafner, D., Lillicrap, T., Ba, J., Norouzi, M.: {Dream to Control}: Learning
  behaviors by latent imagination. In: International Conference on Learning
  Representations (2020)

\bibitem{he2016deep}
He, K., Zhang, X., Ren, S., Sun, J.: Deep residual learning for image
  recognition. In: IEEE Conference on Computer Vision and Pattern Recognition
  (2016)

\bibitem{hespanha2018linear}
Hespanha, J.P.: Linear systems theory. Princeton university press (2018)

\bibitem{hur2023genhpf}
Hur, K., Oh, J., Kim, J., Kim, J., Lee, M.J., Cho, E., Moon, S.E., Kim, Y.H.,
  Atallah, L., Choi, E.: Genhpf: General healthcare predictive framework for
  multi-task multi-source learning. IEEE Journal of Biomedical and Health
  Informatics  (2023)

\bibitem{ishihara2021multi}
Ishihara, K., Kanervisto, A., Miura, J., Hautamaki, V.: Multi-task learning
  with attention for end-to-end autonomous driving. In: IEEE/CVF Conference on
  Computer Vision and Pattern Recognition (2021)

\bibitem{krizhevsky2017imagenet}
Krizhevsky, A., Sutskever, I., Hinton, G.E.: {ImageNet} classification with
  deep convolutional neural networks. Communications of the ACM  (2017)

\bibitem{liang2024pointmamba}
Liang, D., Zhou, X., Wang, X., Zhu, X., Xu, W., Zou, Z., Ye, X., Bai, X.:
  {PointMamba}: A simple state space model for point cloud analysis. arXiv
  preprint arXiv:2402.10739  (2024)

\bibitem{liang2023multi}
Liang, X., Liang, X., Xu, H.: Multi-task perception for autonomous driving. In:
  Autonomous Driving Perception: Fundamentals and Applications, pp. 281--321.
  Springer (2023)

\bibitem{lin2023scale}
Lin, B., Jiang, W., Ye, F., Zhang, Y., Chen, P., Chen, Y.C., Liu, S., Kwok,
  J.T.: Dual-balancing for multi-task learning. arXiv preprint arXiv:2308.12029
   (2023)

\bibitem{linreasonable}
Lin, B., Ye, F., Zhang, Y., Tsang, I.: Reasonable effectiveness of random
  weighting: A litmus test for multi-task learning. Transactions on Machine
  Learning Research  (2022)

\bibitem{liu2021conflict}
Liu, B., Liu, X., Jin, X., Stone, P., Liu, Q.: Conflict-averse gradient descent
  for multi-task learning. In: Neural Information Processing Systems (2021)

\bibitem{liu2024vmamba}
Liu, Y., Tian, Y., Zhao, Y., Yu, H., Xie, L., Wang, Y., Ye, Q., Liu, Y.:
  Vmamba: Visual state space model. arXiv preprint arXiv:2401.10166  (2024)

\bibitem{liu2021swin}
Liu, Z., Lin, Y., Cao, Y., Hu, H., Wei, Y., Zhang, Z., Lin, S., Guo, B.: Swin
  transformer: Hierarchical vision transformer using shifted windows. In:
  IEEE/CVF International Conference on Computer Vision (2021)

\bibitem{loshchilov2018decoupled}
Loshchilov, I., Hutter, F.: Decoupled weight decay regularization. In:
  International Conference on Learning Representations (2019)

\bibitem{ma2024u}
Ma, J., Li, F., Wang, B.: U-mamba: Enhancing long-range dependency for
  biomedical image segmentation. arXiv preprint arXiv:2401.04722  (2024)

\bibitem{maninis2019attentive}
Maninis, K.K., Radosavovic, I., Kokkinos, I.: Attentive single-tasking of
  multiple tasks. In: Computer Vision and Pattern Recognition (2019)

\bibitem{mehta2023long}
Mehta, H., Gupta, A., Cutkosky, A., Neyshabur, B.: Long range language modeling
  via gated state spaces. In: International Conference on Learning
  Representations (2023)

\bibitem{misra2016cross}
Misra, I., Shrivastava, A., Gupta, A., Hebert, M.: Cross-stitch networks for
  multi-task learning. In: IEEE Conference on Computer Vision and Pattern
  Recognition (2016)

\bibitem{sk18}
Sener, O., Koltun, V.: Multi-task learning as multi-objective optimization. In:
  Neural Information Processing Systems (2018)

\bibitem{silberman2012indoor}
Silberman, N., Hoiem, D., Kohli, P., Fergus, R.: Indoor segmentation and
  support inference from rgbd images. In: European Conference on Computer
  Vision (2012)

\bibitem{vandenhende2021multi}
Vandenhende, S., Georgoulis, S., Van~Gansbeke, W., Proesmans, M., Dai, D.,
  Van~Gool, L.: Multi-task learning for dense prediction tasks: A survey. IEEE
  Transactions on Pattern Analysis and Machine Intelligence  \textbf{44}(7),
  3614--3633 (2021)

\bibitem{vandenhende2020mti}
Vandenhende, S., Georgoulis, S., Van~Gool, L.: Mti-net: Multi-scale task
  interaction networks for multi-task learning. In: European Conference on
  Computer Vision (2020)

\bibitem{wang2024graph}
Wang, C., Tsepa, O., Ma, J., Wang, B.: {Graph-Mamba}: Towards long-range graph
  sequence modeling with selective state spaces. arXiv preprint
  arXiv:2402.00789  (2024)

\bibitem{wang2024mambabyte}
Wang, J., Gangavarapu, T., Yan, J.N., Rush, A.M.: {MambaByte}: Token-free
  selective state space model. arXiv preprint arXiv:2401.13660  (2024)

\bibitem{wolf2020transformers}
Wolf, T., Debut, L., Sanh, V., Chaumond, J., Delangue, C., Moi, A., Cistac, P.,
  Rault, T., Louf, R., Funtowicz, M., Davison, J., Shleifer, S., von Platen,
  P., Ma, C., Jernite, Y., Plu, J., Xu, C., Le~Scao, T., Gugger, S., Drame, M.,
  Lhoest, Q., Rush, A.: Transformers: State-of-the-art natural language
  processing. In: Conference on Empirical Methods in Natural Language
  Processing (2020)

\bibitem{xing2024segmamba}
Xing, Z., Ye, T., Yang, Y., Liu, G., Zhu, L.: Segmamba: Long-range sequential
  modeling mamba for 3d medical image segmentation. arXiv preprint
  arXiv:2401.13560  (2024)

\bibitem{xu2018pad}
Xu, D., Ouyang, W., Wang, X., Sebe, N.: Pad-net: Multi-tasks guided
  prediction-and-distillation network for simultaneous depth estimation and
  scene parsing. In: IEEE Conference on Computer Vision and Pattern Recognition
  (2018)

\bibitem{xu2023multi}
Xu, Y., Li, X., Yuan, H., Yang, Y., Zhang, L.: Multi-task learning with
  multi-query transformer for dense prediction. IEEE Transactions on Circuits
  and Systems for Video Technology  \textbf{34}(2),  1228--1240 (2024)

\bibitem{ye2024first}
Ye, F., Lin, B., Cao, X., Zhang, Y., Tsang, I.: A first-order multi-gradient
  algorithm for multi-objective bi-level optimization. arXiv preprint
  arXiv:2401.09257  (2024)

\bibitem{ye2021multi}
Ye, F., Lin, B., Yue, Z., Guo, P., Xiao, Q., Zhang, Y.: Multi-objective meta
  learning. In: Neural Information Processing Systems (2021)

\bibitem{yeadaptive}
Ye, F., Lyu, Y., Wang, X., Zhang, Y., Tsang, I.: Adaptive stochastic gradient
  algorithm for black-box multi-objective learning. In: International
  Conference on Learning Representations (2024)

\bibitem{ye2022inverted}
Ye, H., Xu, D.: Inverted pyramid multi-task transformer for dense scene
  understanding. In: European Conference on Computer Vision (2022)

\bibitem{yu2020gradient}
Yu, T., Kumar, S., Gupta, A., Levine, S., Hausman, K., Finn, C.: Gradient
  surgery for multi-task learning. In: Neural Information Processing Systems
  (2020)

\bibitem{ze2023gnfactor}
Ze, Y., Yan, G., Wu, Y.H., Macaluso, A., Ge, Y., Ye, J., Hansen, N., Li, L.E.,
  Wang, X.: Gnfactor: Multi-task real robot learning with generalizable neural
  feature fields. In: Conference on Robot Learning (2023)

\bibitem{zhang2024point}
Zhang, T., Li, X., Yuan, H., Ji, S., Yan, S.: Point could mamba: Point cloud
  learning via state space model. arXiv preprint arXiv:2403.00762  (2024)

\bibitem{zhang2021survey}
Zhang, Y., Yang, Q.: A survey on multi-task learning. IEEE Transactions on
  Knowledge and Data Engineering  \textbf{34}(12),  5586--5609 (2022)

\bibitem{zhang2019pattern}
Zhang, Z., Cui, Z., Xu, C., Yan, Y., Sebe, N., Yang, J.: Pattern-affinitive
  propagation across depth, surface normal and semantic segmentation. In:
  IEEE/CVF Conference on Computer Vision and Pattern Recognition (2019)

\bibitem{zhou2020pattern}
Zhou, L., Cui, Z., Xu, C., Zhang, Z., Wang, C., Zhang, T., Yang, J.:
  Pattern-structure diffusion for multi-task learning. In: IEEE/CVF Conference
  on Computer Vision and Pattern Recognition (2020)

\bibitem{zhu2024vision}
Zhu, L., Liao, B., Zhang, Q., Wang, X., Liu, W., Wang, X.: {Vision Mamba}:
  Efficient visual representation learning with bidirectional state space
  model. In: International Conference on Machine Learning (2024)

\end{thebibliography}
\end{document}